\pdfoutput=1 
\documentclass[11pt]{article}

\usepackage[preprint]{acl}

\usepackage{times}
\usepackage{latexsym}
\usepackage[T1]{fontenc}
\usepackage[utf8]{inputenc}
\usepackage{microtype}
\usepackage{graphicx}
\usepackage{float}
\usepackage{xurl}
\usepackage{booktabs}
\usepackage{amsmath}
\usepackage{amssymb}
\usepackage{xcolor}

\usepackage[font=small,labelfont=bf,skip=2pt]{caption}

\graphicspath{{figures/}}

\title{Social Chain of Thought: A Multi-Agent Architecture Grounded in Medical Differential Diagnosis Methodology}

\author{Del Coburn \\
  Department of Sociology \\
  University of Toronto \\
  \texttt{del.coburn@utoronto.ca} \And
  Scott Sanner \\
  Department of Mechanical \\
  \& Industrial Engineering \\
  University of Toronto \\
  \texttt{ssanner@mie.utoronto.ca} \And
  Dan Silver \\
  Department of Sociology \\
  University of Toronto \\
  \texttt{dan.silver@utoronto.ca}}

\begin{document}
\maketitle

\begin{abstract}
Medical diagnostic reasoning is a high-impact use case for LLMs that carries significant implications for the health and wellbeing of users. When OpenAI \citeyearpar{openai2026healthcare} reports that more than 5\% of ChatGPT messages globally are healthcare-related, the transparency of these systems becomes a serious design concern. This is especially true for complex cases, where differential diagnosis often requires integrating multiple forms of specialist reasoning. Existing work has proposed multi-agent approaches to medical diagnosis, but it remains unclear when such systems are needed, why they help, and where they outperform monolithic inference. We introduce Social Chain of Thought (SCoT), a multi-round pipeline for medical differential diagnosis that structures multi-agent interaction as a deliberative framework for collaborative LLM reasoning. Evaluating SCoT against single-agent baselines, one-agent pipeline ablations, and best-of-n scaling, we show that its recall advantage is not reproduced by monolithic inference alone. SCoT is most successful in the hardest diagnostic cases, where multiple rounds of specialist conversation help recover ground-truth diagnoses and converge on a higher-recall differential.
\end{abstract}

\section{Introduction}

Over the past three years, multi-agent systems have become increasingly social in how they operate. From the birth of CAMEL \citep{li2023camel} and Smallville \citep{park2023generative}, to complex agentic frameworks seen in frontier applications such as Claude Code \citep{anthropic2025claudecode}, ChatGPT Agent Mode \citep{openai2025chatgptagent}, and Agent Garden by Google \citep{google2025agentgarden}, artificial intelligence (AI) has adopted what could be described as Social Chain of Thought (SCoT). SCoT is more than additional tokens, more parameters, or higher compute; it is the structure that organizes agent interaction to improve the overall reasoning capacity of the system. This approach is especially relevant to differential diagnostic reasoning, where multi-specialist deliberation over complex medical cases helps practitioners arrive at a correct diagnosis.

At its core, SCoT may be viewed as a form of scaling, where multiple agents act as additional reasoning passes for each other, much like the think-again mechanic of its traditional Chain of Thought methods \citep{wei2022cot}. Beyond simply stacking the deck, multi-agent systems come with a built-in solution to the endogeneity problem of single-model self evaluation. As \citet{huang2024selfcorrect} explain, LLM self-evaluation propagates correlation bias inherent to the source of generation. Endogeneity is thus not a problem that can be solved simply by scaling a model or giving a longer time to ``think;'' it must be approached with intentional heterogeneity. This is what SCoT is architecturally designed to mitigate.

SCoT is heterogeneous inference that produces measurable error decorrelation through persona conditioning, offering multiple, specialized perspectives for diagnostic reasoning. We acknowledge that with the same model backend, endogeneity remains an issue, but find the value of incorporating a social dimension into the reasoning process is a more transparent inference. SCoT structures reasoning variance through persona conditioning to simulate cross-model approaches proposed in multi-agent literature \citep{kamoi2024,liang2024divergent,du2023}. This structure traces how the system arrived at a conclusion through a series of interactions, allowing a degree of transparency into performance through deliberative reasoning.

We propose SCoT as a sociology-inspired method to test this form of inference. Our approach extends the Chain of Thought (CoT) dual-inference deployed by \citet{zhou2025} for medical differential diagnoses, using an LLM prompted with a complex symptomatology as a backend for generating a panel of five medical specialists relevant to the case at hand. Using the physician-curated benchmark provided by the authors (Open-XDDx), we deploy SCoT across a range of model families where these dynamically generated, persona-conditioned agents deliberate over the course of a seven round pipeline to arrive at a final differential diagnosis.

\section{Contributions}

We contribute:
\begin{itemize}\itemsep0pt\parsep0pt\topsep0pt\partopsep0pt
\item The SCoT pipeline as a reproducible, open-weight architecture for multi-agent systems design and applied diagnostic reasoning.
\item An open source framework for studying SCoT as support in a clinical setting.
\item Cross-model scaling evidence of multi-agent performance bounds, independent of multi-round effects.
\item SCoT performance metrics, isolated through ablation of multi-round scaling effects.
\item ``Best-of-n'' ablation studies indicating SCoT consistently delivers performance gain in a way that monolithic inference and scaling cannot.
\end{itemize}

\section{Related Work}

\subsection{Multi-agent reasoning frameworks}

This research contributes to the well-established field of multi-agent systems. \citet{yang2025} revisit foundational literature \citep{liang2024divergent,du2023} to test multi-agent systems in comparison to contemporary self-agent scaling methods. SCoT is a parallel method that operationalizes collaborative refinement and diverse exploration of its solution space with no specialist fine-tuning beyond the deliberative structure of its architecture. Additional research by \citet{dhuliawala2023cov} introduces chain-of-verification for multi-agent self-refinement, extended by \citet{zhang2024chainagents} who propose Chain of Agents, to show that multi-agent collaboration is particularly useful in cases of long context reasoning. SCoT enjoys a similar benefit from distributed reasoning compared to centralized monolithic inference.

Evolving orchestration \citep{dang2025evolving} puts agents under the command of a reinforcement learning directive which trains them to adaptively sequence and prioritize behavior towards specific goals. Related to our own work in its design, ChatEval \citep{chan2024chateval} deploys a team of diverse agents specialized to opine on a given text or corpora before voting on which texts may be better. We also see systems such as ReConcile \citep{chen2024reconcile} that separate back-end models entirely to bring multiple LLMs from distinct families into conversation with each other over divisive topics.

Further on the theme of productive, divisive interaction, \citet{guo2024multiagent} detail how feedback from agent interactions allows for refinement as agents are judged by others, leading to critical self-evaluation. \citet{xiong2023ford} introduce their formal debate framework (FORD) as a means to encourage collaborative behaviour in LLMs and agentic systems. The authors explicitly highlight how interactive debate allows agents to ``explore the differences between their own understandings and the conceptualizations of others'' \citep[1]{xiong2023ford}.

While these systems demonstrate that interaction among agents can improve reasoning or evaluation, they do not directly test whether a structured pipeline produces gains beyond repeated inference. \citet{cemri2025} specifically identify heterogeneity in multi-agent systems as distinct from resource scaling, encouraging such systems to maximize this benefit over simply leveraging additional compute.

\subsection{Medical-domain multi-agent diagnostics}

In the medical domain specifically, several multi-agent diagnostic systems have been proposed. \citet{tu2024amie} introduce ``AMIE,'' a system designed to take on roles that might be present in a medical setting, such as a doctor, a patient, a nurse, and a critic to refine its behavior as a patient-facing tool. AMIE is a close relative to our own work in that it optimizes diagnostic reasoning and simulates personas from a similar domain.

\citet{li2025macd} present MACD, a multi-agent system that simulates specialists to organize, curate, debate, and diagnose. This system retains real clinical data as a self-improvement mechanism, deploying it in future cases. While this system runs parallel to our SCoT framework in its application to medical contexts, there are fundamental differences in terms of access to data, agent persona conditioning, and the use of LLM as a judge that distinguish the two approaches.

Related most closely to our work, \citet{chen2025mac} introduce Multi-agent Conversation (MAC), a framework which generates a team of specialist doctors that perform medical consultation. This system is similar to AMIE \citep{tu2024amie} in that it integrates a supervisor agent. Where it primarily differs from our work is in its use of proprietary frontier models toward a single diagnosis, as distinct from the differential diagnoses SCoT is tested on.

\subsection{Monolithic LLMs for medical diagnosis}

Several other approaches using monolithic LLMs for diagnostic reasoning have been tested, with much of the literature concluding they are inadvisable or unhelpful \citep{goh2024jama}. \citet{pal2022medmcqa} evaluate pre-GPT-3 era natural language processing in diagnostic medicine, concluding that even specially trained models show poor performance interpreting medical examination questions. One year later, \citet{chen2023meditron} released MediTron as a high performing open-weight model for medical reasoning. Today, medical students are often consulting DxGPT \citep{dxgpt2025} and ChatGPT Health \citep{chatgpthealth2026} as they learn their craft. The field moves quickly in this domain, with monolithic models largely dominating the space. This is, in part, what makes differential diagnosis such a promising application for SCoT. Strong monolithic baselines exist, but the task they aim to model is fundamentally deliberative.

\section{Research Questions}

\subsection{RQ1}

\paragraph{RQ1-H1: Minimum viable configuration.} For a fixed underlying LLM with equal parameters, how much can the SCoT multiagent approach improve diagnosis recall vs.\ monolithic inference?

Our prediction is informed by \citet{yang2025}, who find that smaller models reap the most from placement in multi-agent systems. While we concur in our prediction, we also propose that there is a minimum floor of capability at which SCoT not only ceases to improve system performance, but detracts from it. This will be measured through model size ablation testing, where parameters are scaled down across runs of all 570 cases in the Open-XDDx benchmark \citep{zhou2025}.

\paragraph{RQ1-H2: Deliberative headroom.} Does SCoT improve the performance of smaller and larger models over monolithic inference?

Pursuing this second line of inquiry, SCoT will again be examined through the lens of \citet{yang2025} to assess the social mechanism that leads to performance gain. We predict that SCoT performance gain will be meaningfully higher for smaller models than for that of its larger counterparts. We term this measure: deliberative headroom. Where a model might be more intrinsically capable in its reasoning, SCoT will offer less benefit as compared to smaller models. This establishes the SCoT performance ceiling.

\subsection{RQ2}

%
\paragraph{RQ2-H3: Multi-agent structure versus multi-round prompting.} Will a single agent achieve better performance through the SCoT pipeline as compared to baseline monolithic inference, and does the multi-agent advantage persist when total LLM computation is held constant?

To isolate SCoT performance gain, we ablate multiple agents from the pipeline, testing if a single agent will achieve increased performance using the pipeline as repeated inference. Our prediction is that SCoT offers specific social advantages for recall that are not directly contingent on additional compute. Single-agent configurations should not see a performance gain of any meaningful significance compared to performance of the multi-agent configuration or its single-agent baseline. Repeated sampling at a matched inference budget should also fall short of the multi-agent configuration, isolating the contribution of agent heterogeneity from raw compute.

Best-of-n tests on Qwen-2.5-32B set a compute equivalence benchmark for test-time scaling, where a single agent engages in repetitive sampling of five prompts-per-round over seven rounds. This experiment isolates the impact of SCoT, testing if the argument for social scaling survives against standard scaling methods. We predict that a single LLM system operating through iterative unstructured inference will not achieve the recall score of multiple agents interacting through SCoT.

\paragraph{RQ2-H4: Difficulty-conditioned consensus formation.} Do harder diagnostic cases rely more on late-stage multi-agent refinement than easier cases?

Following the literature advocating multiple perspectives in LLM simulations \citep{guo2024multiagent,xiong2023ford,cemri2025}, we predict that SCoT consensus dynamics will vary by case difficulty. We use monolithic baseline performance as a proxy for difficulty, and late-stage candidate heterogeneity as its signal in SCoT. Cases where the baseline performs poorly should show more pronounced late-stage consensus formation. This reflects the refinement of proposals as agents use the shared candidate space to recover plausible diagnoses that were not initially supported by the full team. By contrast, in cases where the monolithic baseline performs well, SCoT should converge earlier, benefiting less from additional deliberation when heterogeneity reaches consensus by the independent differential stage (round 3). This hypothesis tests whether SCoT improves aggregate recall by shifting the mechanism of consensus under diagnostic uncertainty.

\subsection{Evaluation}

The Open-XDDx \citep{zhou2025} benchmark dataset contains 570 physician-curated cases, each in the form of a short vignette, ranging from 14 to 285 words. Each vignette contains basic patient demographic information, generally age and sex, and a list of symptoms. See Appendix~\ref{app:openxddx} for a complete list of case types as grouped by specialty.

In line with the evaluation methods used by \citet{zhou2025}, we evaluate SCoT against the Open-XDDx using the standard information retrieval metrics adapted by the authors to differential diagnosis. Each proposed diagnosis is evaluated as a Boolean match against the benchmark ground truth term, where a ground truth diagnosis recovered by the system is counted as a True Positive (TP), a ground truth diagnosis omitted from the final differential is counted as a False Negative (FN), and a proposed diagnosis that does not correspond to a ground truth term is counted as a False Positive (FP). 
We measure standard Recall, Precision, and F1 score against the final ranked differential.



\section{Method \& Design}

\subsection{Persona conditioning and specialist generation}

Before the first round of the pipeline, the backend LLM is provided with the clinical vignette of a single case from the Open-XDDx \citep{zhou2025} dataset. It generates a team of five agents and assigns each a distinct persona as a medical specialist deemed relevant to the case. Every specialist team is composed of a unique configuration depending on how the LLM has conditioned their personas. This feature is benchmark-agnostic and will generate a panel of specialist agents for any clinical input it receives.

For ecological validity, refinement and consensus stages of the SCoT pipeline are informed by clinical Delphi methods. \citet{graham2003delphi} use an expert panel to establish diagnostic criteria, respond to group-level feedback, and build consensus across iterative rounds. We adapt this process for SCoT differential diagnosis, where agents independently propose candidate diagnoses, deliberate in a shared diagnostic space, and synthesize a final ranked differential through structured voting.

\subsection{The SCoT pipeline}

\textbf{Round 1: Specialized Ranking} is the first round after a team of specialists has been generated by the backend LLM. Each agent is instructed to assess the case from the perspective of their specialty and to provide a self-assessment of their relevance to the case in question.

\textbf{Round 2: Symptom Management} is inspired by ``AMIE'' \citep{tu2024amie}, and models the triage process in a medical setting. The intention of this round is to reflect a realistic healthcare environment where a patient would be treated for certain obvious or life-threatening conditions prior to seeing a specialist. All agents perform this triaging process.

\textbf{Round 3: Team Independent Differentials} instructs agents to consider the case from a strictly diagnostic angle, each one delivering their independent assessment as a list of most likely diagnoses from the perspective of their specialty.

\textbf{Round 4: Master List} compiles the independent differentials contributed by each agent into an unranked and deduplicated list to ensure each diagnosis only appears once when provided to each agent in round five for deliberation.

\textbf{Round 5: Refinement} is a set of three sub-rounds that draws on agent debate frameworks by \citet{yang2025}. In the first, agents state their initial positions in relation to each diagnosis on the master list. They explicitly state Support, with evidence (the agent provides evidence for why they support the diagnosis), Challenge, with reasoning (agents must provide reasoning for their challenge), and Neutral for diagnoses they feel require more information. The second sub-round requires agents to challenge a diagnosis provided by another agent, to propose a diagnosis they feel another agent might have missed, to ask another agent a specific question about their reasoning, or to cite evidence that contradicts or complicates the diagnosis of another agent. In the final sub-round, each agent will address each challenge against them directly, defend their position if they have strong counter-evidence, update their diagnoses if the challenge changed their reasoning, and to state clearly which positions they maintain and which positions they changed.

\textbf{Round 6: Voting} takes place after agents are provided the final diagnoses coming out of the refinement round. Each agent ranks each diagnosis from most plausible to least plausible. These scores are then aggregated and internally weighted by a combination of three response-derived heuristic components (diagnostic insight, breadth, and actionability) before Borda counting synthesizes a final differential diagnosis (see Appendix~\ref{app:credibility} for component details).

\textbf{Round 7: Can't Miss} functions as a final safety check for agents to review the case and propose any critical or life-threatening diagnoses that may have been missed. Agents are prompted to identify diagnoses that require extreme urgency. This round was initially idealized for any real-life deployment in a clinical setting where these diagnoses would be raised as those considered most critical or life-threatening.

\subsection{Backend models}

Our experimental design largely focuses on open-weight models, with closed-weight counterparts tested to establish frontier viability. All models, with the exception of Claude Haiku 4.5 and Sonnet 4.6, have at least one full run of the Open-XDDx through both single-agent baseline and SCoT conditions. Claude models perform 100 runs of each condition as an exploratory comparison. Qwen-2.5 was used for size and multi-agent ablation testing.

Temperature settings for backend models are set to a default of 0.3 and 0.7. Each team of specialists has persona conditioned agents split into two groups, innovative and conservative. Innovative agents operate with a temperature setting of 0.7, while conservative agents operate with a temperature setting of 0.3. Each team is composed of two innovative specialists and three conservative specialists. We perform ablation testing to determine the effect of temperature on our experiment, as detailed in Section~\ref{sec:ablation} and Appendix~\ref{app:ablations}.

\smallskip
\noindent Backend models tested:
\begin{itemize}\itemsep0pt\parsep0pt\topsep0pt\partopsep0pt
\item Qwen-2.5: 1.5B, 3B, 32B (GPTQ-Int4) \citep{qwen2025}
\item Gemma-4-MoE-26B-A4B-it-AWQ-4bit \citep{gemma2026}
\item Gemma-4-31B-it-AWQ \citep{gemma2026}
\item GPT-5-4 nano \citep{openai2025gpt5}
\item Claude Haiku 4.5 \citep{anthropic2025haiku45}
\item Claude Sonnet 4.6 \citep{anthropic2025sonnet46}
\end{itemize}

\subsection{Ablation Testing and Benchmarking}\label{sec:ablation}

Each model tests against a single-agent baseline prompted with a case and instructed to provide a ranked differential diagnosis, over 570 cases without the SCoT pipeline. We also test against \citet{zhou2025}, who report 53\% recall using dual-inference with ChatGPT-4o. The single-agent baseline is used as a monolithic inference benchmark for SCoT, while \citet{zhou2025} are considered an external comparative benchmark. In an extension of the baseline, we run a single agent through the SCoT pipeline against all 570 cases of the Open-XDDx \citep{zhou2025}. The agent is generated and persona conditioned as a specialist by its backend LLM before proceeding through the seven rounds described in Section~5.2.

Multiple ablation studies test our hypotheses. Five temperature sweeps on Qwen-2.5: 32B and Gemma-4-MoE-26B were run across all 570 cases showing temperature has little effect on our experiments, and include further details in Appendix~\ref{app:ablations}. Best-of-n experiments are used to determine whether SCoT performance gains are matched by repeated sampling methods. This entails prompting the backend LLM with each case n number of times and instructing it to provide a differential diagnosis. Drawing on research by \citet{brown2024monkeys}, we use the best available samples to level the playing field against the implicit compute of five agents over seven rounds, giving Qwen-2.5-32B 35 inference passes over seven rounds for each case, aggregated by majority vote. We run additional tests of n at 35 that use Qwen-2.5-32B as a judge to emulate the reward-model outlined by \citet{huang2025bestofn}. We ensure these tests are not reward hacking by limiting repeated inference to comparable compute.

We run an additional test of SCoT with three agents instead of five to determine the impact of the number of agents. This setup is identical to SCoT runs with five agents, with the only meaningful variable modification being the number of agents in the simulation (one innovative, two conservative). We include results from all ablation testing in Appendix~\ref{app:ablations}.

\section{Results}

Results are reported as aggregates across model families. For full panel specifics and experimental outcomes, see Appendix~\ref{app:cross}. In the following subsections, we report results according to their corresponding research question and hypothesis from Section~4.

\subsection{RQ1 -- H1-H2: Minimum viability and deliberative headroom}

Across models, SCoT improves recall by four to twelve percentage points on every backend with the exception of Qwen-2.5-1.5B. H1 is supported, with the small surprise of degraded performance in the 1.5B model, which falls by 3.7 percentage points relative to its single-agent baseline. The Qwen-2.5 size tests indicate that the SCoT viability threshold falls between 1.5B and 3B parameters: Qwen-2.5-1.5B degrades under SCoT, while Qwen-2.5-3B gains 11.8 recall points over its single-agent baseline.

\begin{figure}[!ht]
\centering
\includegraphics[width=\columnwidth]{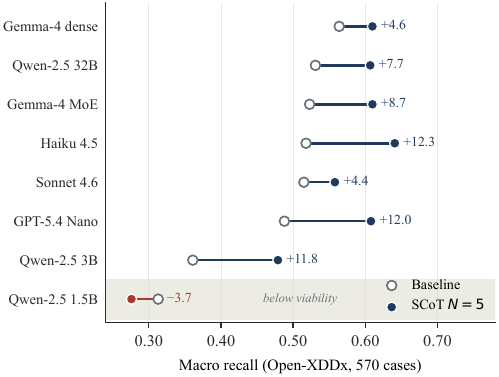}
\caption{Cross-model deliberative headroom: baseline and SCoT ($N{=}5$) macro recall on Open-XDDx, with per-model deltas (pp). Claude and Gemma-dense use 100 exploratory cases; all others use the full 570 (Appendix Table~\ref{tab:cross-panel}).}
\label{fig:headroom}
\end{figure}

We predicted that the performance gain from SCoT will be greater in magnitude for smaller models, with this boost decreasing inversely with model size. This hypothesis is supported across model families, with evidence for what we earlier termed deliberative headroom. Models with lower-performing single-agent baselines have more recoverable diagnostic capacity than models already closer to the SCoT performance ceiling. Figure~\ref{fig:headroom} shows SCoT and baseline performance across model sizes and families.

\subsection{RQ2 -- H3: Multi-specialist structure versus multi-round prompting}

Our second research question asked whether SCoT performance gain was socially induced or if this was simply a matter of extra compute through structured test-time scaling. The main hypothesis predicted that a single agent in a SCoT-configured pipeline would not see a performance gain of any meaningful significance in relation to its simple baseline. This finds support, with some qualification.

\begin{figure}[!ht]
\centering
\includegraphics[width=\columnwidth]{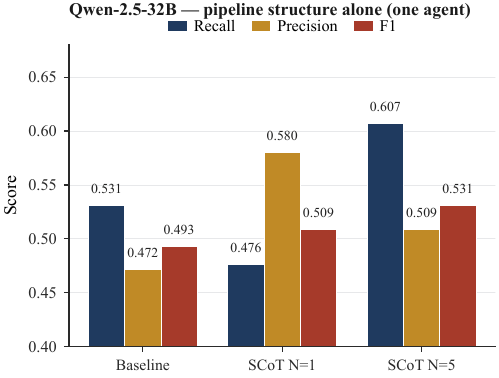}
\caption{Single-agent SCoT compared to baseline and multi-agent SCoT (Qwen-2.5-32B, 570 cases). Pipeline structure alone trades recall for precision; the multi-agent configuration recovers and extends recall.}
\label{fig:h3}
\end{figure}

As detailed in Section~\ref{sec:ablation}, we ran SCoT with a single-agent through all 570 cases of the Open-XDDx. In this setup, we found that recall degraded relative to its single-agent baseline (no SCoT), losing 5.44 percentage points. The configuration did, however, improve precision by 10.74 percentage points, suggesting that the scaffold of the pipeline functions as a precision-positive filter for a single agent, but as a recall-positive broadening mechanism for multiple agents (Figure~\ref{fig:h3}).

This finding is largely a structural feature of the pipeline that pushes the single-agent to refine its reasoning over each round. Proposed diagnoses are filtered through the voting mechanism, where its differential gains precision as it is stripped of noise that would otherwise be captured as false positives. Performance before the voting round sees the single-agent trade precision for only a marginal recall gain: precision falls by 11.1 percentage points while recall increases by 4.2 percentage points. These findings support SCoT pipeline as a mechanism for refinement; in the absence of heterogeneity, a single agent will not generate the breadth of diagnostic candidates. Single agents do not tend to add to their initial proposed differential, instead using the rounds to shave off potential candidates from the final output.

\begin{figure}[!ht]
\centering
\includegraphics[width=\columnwidth]{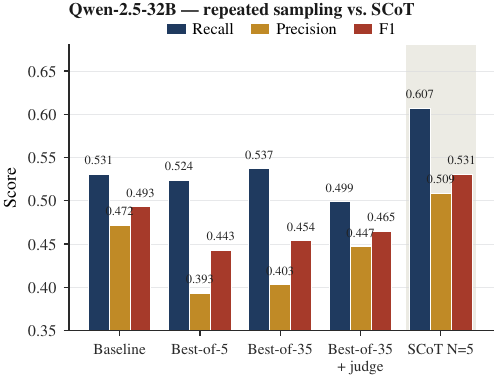}
\caption{Best-of-$n$ repeated sampling compared to SCoT $N{=}5$ (Qwen-2.5-32B). Repeated sampling, including with a judge, does not match SCoT recall.}
\label{fig:bestofn}
\end{figure}

Our best-of-n tests produced results that further suggest SCoT acts as a refinement mechanism. Where the single-agent SCoT-configured pipeline showed an increase in precision, repeated sampling does not reproduce this effect (Figure~\ref{fig:bestofn}). Across cases, precision decreased relative to the single-agent baseline. With 35 samples, repeated inference achieved a recall of 0.537, precision of 0.403, and an F1 of 0.454. Adding a judge is negligible, decreasing recall to 0.499, while increasing precision to 0.447, and F1 to 0.465. The outcome of H3 suggests that structure matters most for the heterogeneity of multiple agents, where they can refine diversity into a more robust differential.

\subsection{RQ2 -- H4: Difficulty conditioned consensus}

The fourth hypothesis predicted that SCoT would be most useful in cases where monolithic inference performs poorly. If the value of SCoT comes from structured diagnostic heterogeneity, then difficult cases should benefit more from late-stage refinement, while easier cases should converge earlier and gain less from additional deliberation. This hypothesis is supported, with some qualification.

\begin{figure}[!ht]
\centering
\includegraphics[width=\columnwidth]{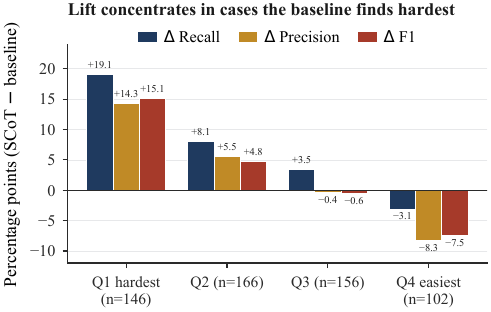}
\caption{Per-quartile change in recall, precision, and F1 (SCoT $-$ baseline), with quartiles defined by baseline difficulty. Lift concentrates in Q1 and flips sign by Q4.}
\label{fig:quartile}
\end{figure}

Our results show that SCoT has its most significant impact in cases where the baseline performs worst, boosting aggregate F1 score by 15.1 percentage points (0.212 -- 0.363), for a relative lift of 71.4\% (Figure~\ref{fig:quartile}). Conversely, in the upper quartile of easiest cases, defined as those where the single-agent baseline performed best, SCoT shows an 8.3 percentage point decrease in precision. This indicates that SCoT helps most in cases where baseline inference is uncertain or incomplete, while in the easiest cases it is almost a matter of overthinking that hurts performance.

Across Qwen-2.5-3B, Qwen-2.5-32B, Gemma-4-MoE, and GPT-5.4 nano, single-agent baselines completely failed on an average of 89 cases, recovering zero ground-truth diagnoses in those cases. Across these baseline-failure cases, the corresponding SCoT runs recovered an average of 16.3\% of ground-truth terms and recovered at least one ground-truth diagnosis in 53.9\% of cases, rescuing 48 of the 89 complete failures.

Looking at the full aggregate, SCoT captures a mean total of 703 ground-truth terms, with 11\% of them entering consensus during the refinement round. In the 89 baseline-failure cases, this proportion rises to 35\%, with SCoT recovering at least one ground-truth term in 48 of those 89 cases. This suggests that refinement becomes especially important when monolithic inference fails completely.

\begin{figure}[!ht]
\centering
\includegraphics[width=\columnwidth]{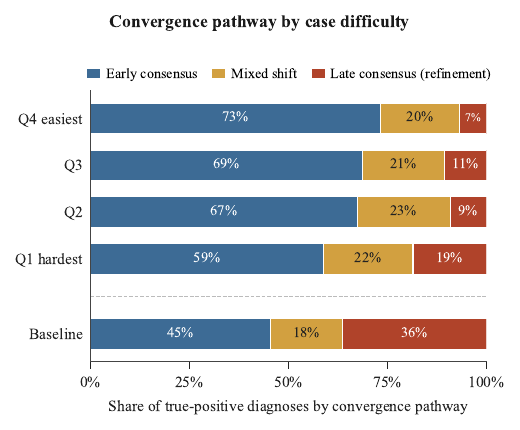}
\caption{SCoT convergence pathway by case difficulty (Qwen-2.5-32B). Quartiles defined by baseline F1. ``Baseline'' row is the subset where the monolithic baseline recovered zero ground-truth diagnoses.}
\label{fig:consensus}
\end{figure}

This early-versus-late convergence pattern emerges across difficulty. For Qwen-2.5-32B, the share of true-positive diagnoses entering through late refinement grows with case difficulty: 7 percent in the easiest quartile, 19 percent in the hardest, and 36 percent on the subset where the baseline returns zero (Figure~\ref{fig:consensus}). Cases that rely on late-round refinement are the same cases the baseline performs worst on. This is the empirical signature of social scaling, where a meaningful portion of SCoT performance gain specifically comes from agents revising positions in response to one another, not from any single agent producing the right answer on the first attempt.

These analyses clarify the viability threshold identified in H1. Qwen-2.5-1.5B initially appeared to mark the lower bound of SCoT viability because its aggregate performance degraded relative to its single-agent baseline. Investigation reveals that although Qwen-2.5-1.5B recovers 70 ground-truth terms through SCoT in cases where its baseline completely fails, it also creates more failures than it resolves. SCoT fails entirely in 159 cases where the corresponding baseline recovers 122 ground-truth terms. Below the viability threshold, multi-agent heterogeneity has not reached a net positive effect. The 1.5B model lacks the underlying knowledge to push back on weak proposals from other agents, placing it below the threshold at which multiple perspectives become productive rather than destabilizing. At 1.5 billion parameters, this is the empirical lower bound of social inference in differential diagnostic reasoning.

\section{Discussion}

The theoretical limits of LLM performance \citep{mohsin2025} are defined by the capacity to operate within a given domain. In that of diagnostic reasoning, social scaling is contingent on the capability for a model to collaborate effectively, where compute only matters insofar as heterogeneity is net positive in its product. Diagnostic systems incapable of supporting a balance of heterogeneity and consensus will converge on the wrong conclusions before the candidate space has broadened to include the right ones. Diversity of perspective in differential diagnostics is effective only insofar as the system is structured around a capable knowledge base in the first place. Diversity can then expose the right answer while consensus mechanisms filter out the noise of speculation.

This same structure filters diagnoses to improve precision for a single agent, but reduces the capacity of the agent to think past its first opinion, also seen in our best-of-n tests where the model falls into a self-affirming feedback loop. The single-agent SCoT tests show higher precision when the agent has a structure to bound it, at the cost of a narrower recall across the board. Narrowing of diagnostic candidates is complemented by heterogeneity in the multi-agent group, whereby broadening the deliberative space makes interaction productive by turning consensus into a mechanism for selecting the right diagnoses out of many possibilities.

This is especially true in late-stage reasoning. What we have identified as case difficulty manifests as poor performance in single-agent approaches, where the complexity of a case is not always captured by more compute or repeated inference. When challenged by the complexity of a difficult case, SCoT pushes the system to broaden its domain of inference. It is a method that captures what iterative optimization leaves on the table. Considering the implications of differential diagnostic reasoning, this is a feature that should not be left out of system design.

\raggedbottom
\section{Conclusion}

SCoT demonstrates that performance in medical diagnostic reasoning can be increased through a social architecture. We have proposed a structure that pushes the system to explore pathways in its knowledge that may be left untouched by standard inference. SCoT performance gain comes from social scaling that broadens the space of active inference to come to a conclusion through deliberation. This is a complementary approach to iterative scaling that bears consequence on diagnostic systems with public-facing application. When someone turns to an LLM for health advice, a narrow path of reasoning puts that individual at risk of a narrow answer. Health is complex. When we design systems that the public comes to rely on for insight into their condition, are we designing them with the capacity for complex reasoning, or are we considering the use-case solely through the lens of optimization? SCoT complements systems design by making use of the inherent capacity of an LLM for deliberative reasoning. When monolithic inference struggles, this method for social scaling shows promise in pushing that model to consider what it might not in solitude. Future work in LLM social scaling is well positioned to explore this dynamic further, and should look to understand conditions of performance in domains beyond diagnostic reasoning.

\section*{Limitations}

Our demonstration of SCoT is not exhaustive. There remain additional configurations and mechanisms for future work to examine. Comparison with frontier models is exploratory, and what has been presented is restricted by the cost of proprietary API usage. There is also room for additional testing with open-weight models, especially as they become more accessible and capable on consumer grade hardware. Open-weight models are not necessarily apples to apples in terms of parameters and quantization, and what we test attempts to make up for in breadth what it does not test in terms of direct technical specifications. Our choice of Qwen-2.5-32B as the feature model is informed by \citet{zhou2025}, given the comparative capability with GPT-4 series that the authors used in their study using the Open-XDDx.

Furthermore, while the dataset is valuable, we recognize that it is only 570 cases that are organized in a specific format. We cannot make a claim that our framework will behave in the same way with data that is organized and presented in a different structured manner (i.e., a patient chart).

Finally, we recognize that executing the pipeline may not be an accessible option given the compute required to run viable LLMs. While consumer grade hardware is becoming increasingly capable of deploying language models, it is not a reasonable expectation that such options are available to all. Consideration of open source and proprietary technology as inherently gated by access to hardware underpins development of this project.

\section*{Ethical Considerations}

Real-life application involves sensitive data. Open-weight deployment mitigates this, but data provenance should be a major concern for any use-case beyond experimentation. On the point of clinical deployment: a real use-case would be in support of diagnostic decision making, not an authoritative source of truth. This system does not intend to replace human-to-human health care delivery.

This research does not pretend its domain is bias free, and acknowledges that healthcare itself has bias in the development of its practices and innovations. Given that such knowledge and practices could propagate to training data, we treat all outputs as possible replication of these issues. Biases within models may have a tendency to overlook certain diagnoses while privileging others, projecting stereotypes they have been trained on.

\bibliography{combined}

\section*{Use of AI Assistance}

Claude Code (Anthropic) was used for coding assistance during pipeline implementation and data analysis. The authors take full responsibility for the original conception of this research, all design decisions, the interpretation of results, and the written content of this paper.

\clearpage
\appendix

\section{Open-XDDx}\label{app:openxddx}

\begin{figure}[H]
\centering
\includegraphics[width=\columnwidth]{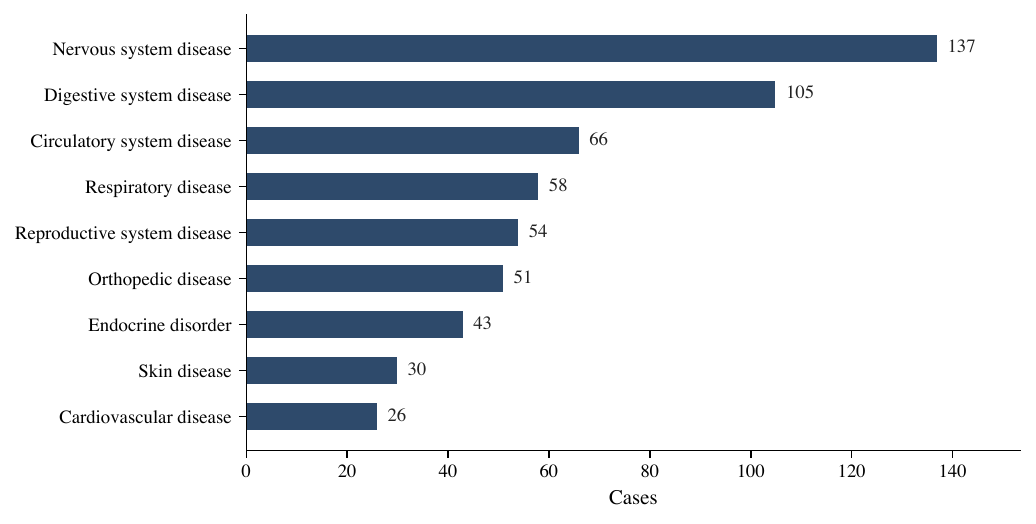}
\caption{Specialty grouping across the 570 Open-XDDx cases.}
\label{fig:specialty}
\end{figure}

\citet{zhou2025} curated the Open-XDDx benchmark used throughout this work. The 570 physician-labelled vignettes group into nine specialty buckets (Figure~\ref{fig:specialty}); a summary of dataset scale and ground-truth term counts per case is given in Table~\ref{tab:dataset}.

\begin{table}[H]
\centering
\caption{Open-XDDx dataset summary: vignette length, ground-truth diagnosis term counts per case, and specialty bucket totals.}
\label{tab:dataset}
\includegraphics[width=\columnwidth]{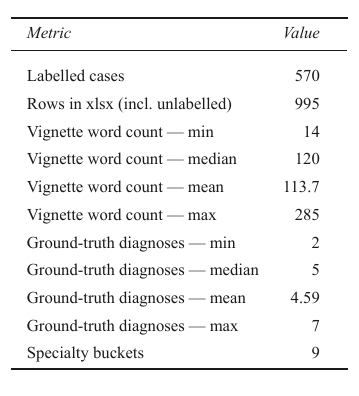}
\end{table}

\section{Cross Model Results}\label{app:cross}

\begin{figure}[H]
\centering
\includegraphics[width=\columnwidth]{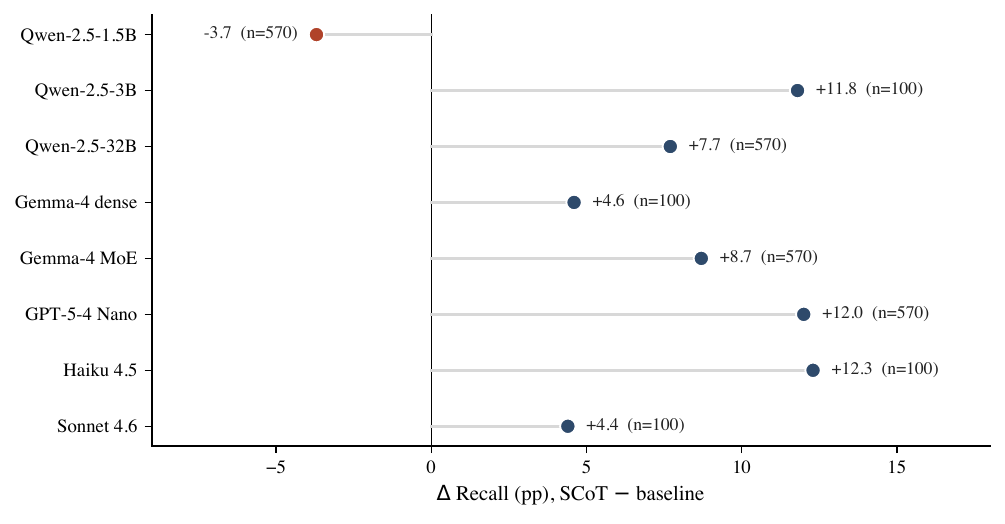}
\caption{Cross-model recall delta (SCoT $-$ baseline) per backend.}
\label{fig:cross-delta}
\end{figure}

The full panel of single-agent and multi-agent SCoT scores across all backend models is given in Table~\ref{tab:cross-panel}. Figure~\ref{fig:cross-delta} renders the panel as a per-model recall delta.

\begin{table}[H]
\centering
\caption{Cross-model panel: macro recall, precision, and F1 under monolithic baseline and SCoT $N{=}5$ conditions, with per-metric deltas.}
\label{tab:cross-panel}
\includegraphics[width=\columnwidth]{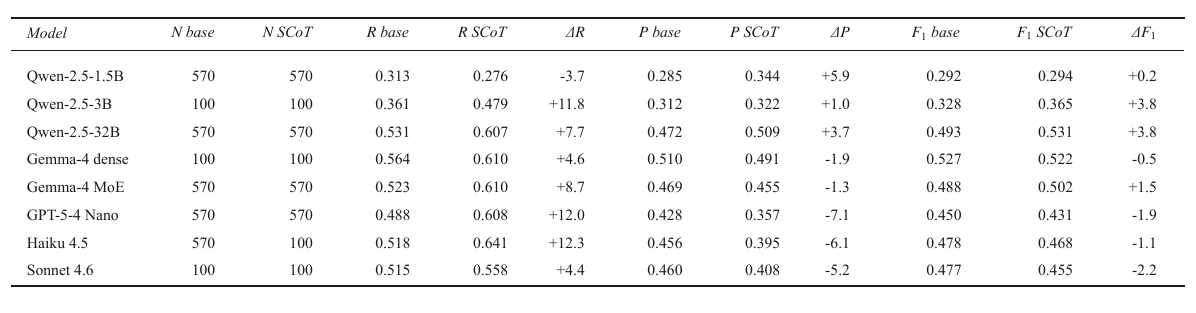}
\end{table}

\section{Ablation Studies}\label{app:ablations}

\begin{figure}[H]
\centering
\includegraphics[width=\columnwidth]{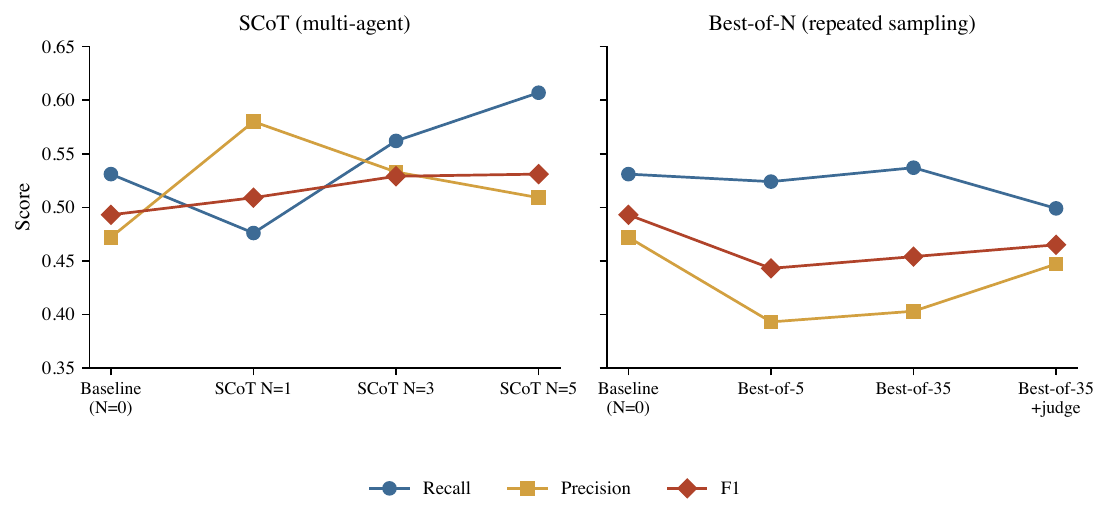}
\caption{Specialist dose-response across the full agent-count and best-of-$n$ sweep.}
\label{fig:dose-curve}
\end{figure}

Full per-condition scores for the agent-count ablation and the best-of-n compute-equivalence runs on Qwen-2.5-32B are given in Table~\ref{tab:dose}, with side-by-side visualisation in Figure~\ref{fig:dose-curve}. Temperature ablations across Qwen-2.5-32B and Gemma-4-MoE-26B are reported in Table~\ref{tab:temp-sweep}.

\begin{table}[H]
\centering
\caption{Performance per condition (Qwen-2.5-32B, 570 cases): baseline, SCoT $N{=}1,3,5$, and best-of-$n$ with and without judge.}
\label{tab:dose}
\includegraphics[width=\columnwidth]{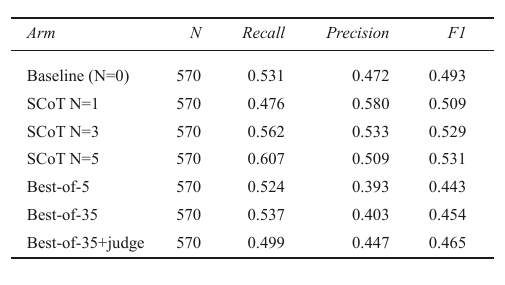}
\end{table}

\begin{table}[H]
\centering
\caption{Cross-model temperature ablations.}
\label{tab:temp-sweep}
\includegraphics[width=\columnwidth]{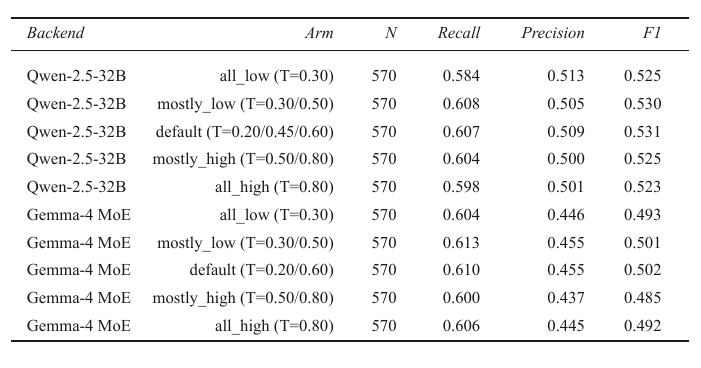}
\end{table}

\section{Error-Correlation Analysis}\label{app:errorcorr}

\begin{figure}[H]
\centering
\includegraphics[width=\columnwidth]{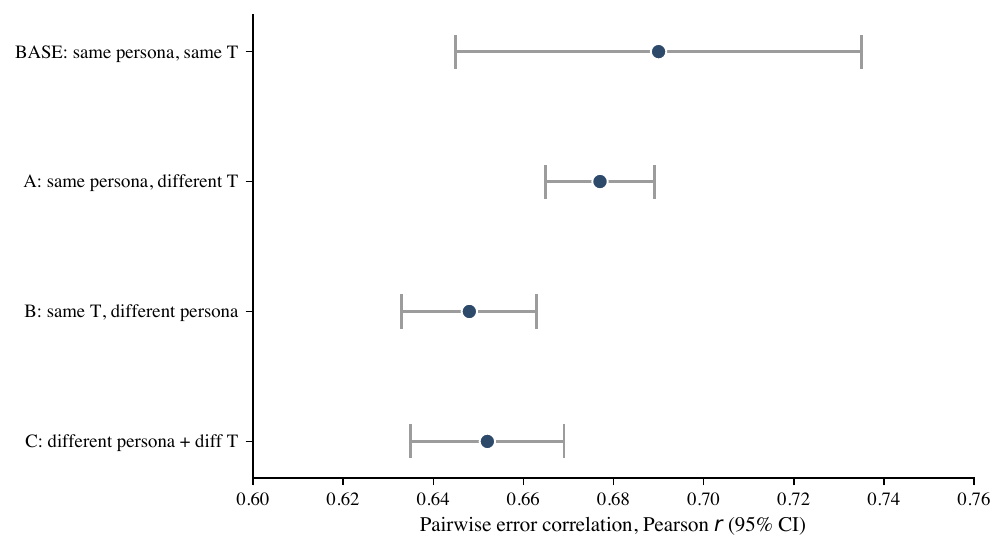}
\caption{Temperature + persona conditioning correlation variance.}
\label{fig:temp-persona-corr-fig}
\end{figure}

To separate temperature heterogeneity from persona heterogeneity, we measured per-case error correlation between agents under matched-temperature / matched-persona conditions (Table~\ref{tab:temp-persona-corr}) and under matched-temperature / different-persona conditions (Table~\ref{tab:persona-corr}). Figure~\ref{fig:temp-persona-corr-fig} visualises the resulting correlation variance.

\begin{table}[H]
\centering
\caption{Temperature + persona conditioning error correlation (Qwen-2.5-32B).}
\label{tab:temp-persona-corr}
\includegraphics[width=\columnwidth]{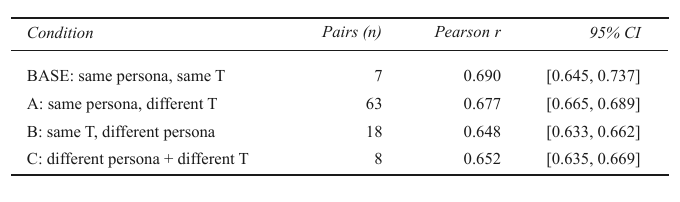}
\end{table}

\begin{table}[H]
\centering
\caption{Persona-only error correlation (temperature held fixed).}
\label{tab:persona-corr}
\includegraphics[width=\columnwidth]{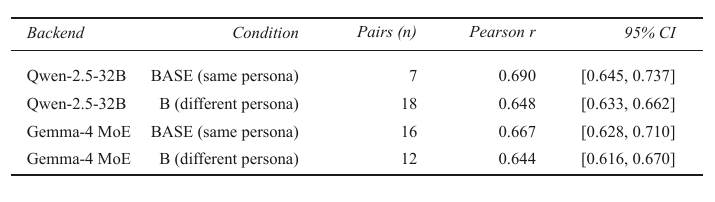}
\end{table}

\section{Deterministic Noise Injection}\label{app:dni}

For robustness testing, we deploy deterministic noise injection to simulate the effect of malicious agents or information. This allows us to see whether the pipeline is able to perform when one of its parts is working against it, or if it requires all agents to behave a certain way. Using both Qwen-2.5-32B and Gemma-4-MoE backends, we instantiate the SCoT pipeline with a deterministic agent that is not backended by an LLM that suggests incorrect diagnoses and votes for itself in every case. These experiments are run on a sample of 200 cases, graded to add an additional deterministic agent for each 200-case run.

\section{Credibility Weighting Prior to Borda Counting}\label{app:credibility}

\subsection{Component scores}
For each specialist, three response-derived scores are computed over the contributions of that specialist across all deliberative rounds. Let $n$ denote the number of distinct diagnoses the specialist proposes across rounds, and let $\bar{e}$ denote the average number of evidence items the specialist supplies per proposed diagnosis.

\emph{Diagnostic insight} rewards the joint production of multiple distinct diagnoses and the depth of evidence supporting them:
\[
\text{Insight} = 3n + 2\bar{e}
\]

\emph{Breadth} rewards the range of distinct diagnoses considered, independent of the evidence attached to each:
\[
\text{Breadth} = 4n
\]

\emph{Actionability} is a binary indicator of whether the leading diagnosis from the specialist is supported by at least two evidence items, awarding $20$ points when satisfied and $0$ otherwise. The threshold operationalizes the requirement that a top-ranked hypothesis be substantively defended rather than asserted.

\subsection{Base diagnostic excellence score}
The three components are combined into a weighted sum:
\begin{multline*}
\text{DES}_\text{base} = 2.5 \cdot \text{Insight} \\
+ 1.5 \cdot \text{Breadth} + 1.0 \cdot \text{Actionability}
\end{multline*}

The coefficients give precedence to evidence-anchored diagnostic quality (insight) over raw differential breadth and over actionability of the leading hypothesis. The 2.5 / 1.5 / 1.0 ratio encodes the intuition that a specialist who proposes well-supported alternatives is more valuable to the consensus than one who lists many candidates without evidence, who is in turn more valuable than one who commits prematurely to a single well-evidenced answer.

\subsection{Professional valence multiplier}
The base score is then modulated by a discourse-quality multiplier $v \in \{0.6, 0.8, 1.0, 1.2\}$ derived from two signals over the combined response text of the specialist: the count of elevating indicators (twelve discourse markers including \textit{evidence}, \textit{guidelines}, \textit{alternative}, \textit{challenge}, \textit{pathophysiology}, \textit{mechanism}) and the total response length. Specialists with four or more elevating indicators and substantive response length receive $v = 1.2$; those meeting weaker thresholds receive $v \in \{1.0, 0.8\}$; minimal contributions receive $v = 0.6$. The valence multiplier formalizes that influence in a deliberative panel is earned through substantive engagement, not asserted by role:
\[
\text{DES}_\text{final} = v \cdot \text{DES}_\text{base}
\]

\subsection{Credibility-weighted Borda aggregation}
In the voting round each specialist $a$ in the active panel submits a preferential ranking $\text{rankings}_a$ over the consolidated diagnosis list. Each ranked entry $r \in \text{rankings}_a$ has a position $\text{pos}(r) \in \{1, 2, 3\}$ for first, second, or third choice, and corresponds to a single candidate diagnosis. Borda points are assigned as $4 - \text{pos}(r)$, yielding 3, 2, and 1 points for first, second, and third choice respectively.

The contribution from specialist $a$ is then scaled by a capped credibility weight:
\[
w_a = \min\!\left(\text{DES}_\text{final}^{(a)},\; 2 \cdot \tilde{c}\right)
\]
where $\text{DES}_\text{final}^{(a)}$ is the final diagnostic excellence score for specialist $a$ (defined earlier in this section) and $\tilde{c}$ is the median of $\text{DES}_\text{final}$ across the active specialist panel. The cap at twice the median prevents any single high-credibility specialist from dominating the aggregate, preserving the deliberative character of the panel while still rewarding substantive contribution.

The final score for a candidate diagnosis $d$ is:
\[
S(d) = \sum_{a \in \text{panel}} \sum_{r \in \text{rankings}_a} \mathbb{1}[r = d] \cdot (4 - \text{pos}(r)) \cdot w_a
\]
where $\mathbb{1}[\cdot]$ is the indicator function returning $1$ when its argument holds and $0$ otherwise. The diagnosis maximizing $S(d)$ is the leading differential of the panel; the full ranking by $S$ is reported as the consensus differential diagnosis.

\end{document}